# SECOE: Alleviating Sensors Failure in Machine Learning-Coupled IoT Systems

Yousef AlShehri
*School of Computing*
*University of Georgia*
Athens, United States
yousef.alshehri@uga.edu

Lakshmish Ramaswamy
*School of Computing*
*University of Georgia*
Athens, United States
laksmr@uga.edu

*Abstract*—Machine learning (ML) applications continue to revolutionize many domains. In recent years, there has been considerable research interest in building novel ML applications for a variety of Internet of Things (IoT) domains, such as precision agriculture, smart cities, and smart manufacturing. IoT domains are characterized by continuous streams of data originating from diverse, geographically distributed sensors, and they often require a real-time or semi-real-time response. IoT characteristics pose several fundamental challenges to designing and implementing effective ML applications. Sensor/network failures that result in data stream interruptions is one such challenge. Unfortunately, the performance of many ML applications quickly degrades when faced with data incompleteness. Current techniques to handle data incompleteness are based upon data imputation – i.e., they try to fill-in missing data. Unfortunately, these techniques may fail, especially when multiple sensors' data streams become concurrently unavailable (due to simultaneous sensor failures). With the aim of building robust IoT-coupled ML applications, this paper proposes SECOE – a unique, proactive approach for alleviating potentially simultaneous sensor failures. The fundamental idea behind SECOE is to create a carefully chosen ensemble of ML models in which each model is trained assuming a set of failed sensors (i.e., the training set omits corresponding values). SECOE includes a novel technique to minimize the number of models in the ensemble by harnessing the correlations among sensors. We demonstrate the efficacy of the SECOE approach through a series of experiments involving three distinct datasets. The experimental findings reveal that SECOE effectively preserves prediction accuracy in the presence of sensor failures.

## I. Introduction

The Internet of Things (IoT) paradigm is transforming many domains of human endeavor, such as agriculture, healthcare, physical infrastructure management, manufacturing, and transportation. At its core, IoT is a collection of Things "objects," each embedded with a sensor, Radio-Frequency Identification (RFID) tag, or actuator, and other software or hardware technologies. Things produce, communicate, and exchange data with one another, connecting a variety of objects such as home appliances, vehicles, and farm equipment to the Internet [1]. IoT has endowed diverse, geographically distributed sensors to seamlessly interact and exchange data over the Internet, which in turn has led to tremendous proliferation of IoT-driven technologies in various fields.

IoT domains are characterized by high-velocity data streams originating from diverse devices. This provides the opportunity to build innovative Big Data applications by constructing models that harness historical and current data, thereby gaining valuable insights that facilitate better decision-making. In this context, there is significant research interest in applying machine learning (ML) techniques for IoT domains. Indeed, many ML applications have already been built for IoT data. As a prominent example of an ML-coupled IoT application, Google uses ML to improve the energy efficiency of its data centers by using data collected from sensors within the data centers, thereby reducing the cost of cooling energy by 40% [2]. Focusing on achieving the ultimate business objective (e.g., reducing power consumption), ML models could be built using supervised ML methods in which a model, using the labeled historical IoT data, learns to map the target goal $y$ to the independent variables as an input $x$ [3]. This model can automatically label a new test sample from sensors to a corresponding prediction in real-time.

However, IoT environments possess several unique characteristics that pose significant challenges for building effective ML-based applications. Some of these include: (1) The need for integrative analytics on heterogeneous data streams; (2) Stringent response time requirements (i.e., many applications are real/semi-real time); (3) Harsh operating environments that are not only resource-constrained but also dynamic; (4) Susceptibility to drastic contextual changes; and (5) Sudden and unexpected interruption of data streams due to IoT-sensor or communication failures resulting in inference-time data incompleteness.

This paper addresses the last of the above-listed challenges, namely, inference-time data incompleteness in ML-coupled IoT systems. Inference time data incompleteness occurs due to data stream interruptions that are commonly caused by two reasons. First, sensor devices can fail either due to electronic malfunctioning or battery outage. Second, communication may fail due to wired/wireless network disruptions. In face of such failures, ML applications are required to make inferences based on incomplete/partial data. As a simple example, consider a hypothetical ML application that continuously determines operational safety of a manufacturing plant based on periodic readings from 4 distinct sensors (temperature, pressure, voltage, and humidity sensors) located at various locations of the manufacturing plant. If two of these sensors fail and stop sending data, the ML application has to continue making inferences based on partial data (i.e., reading from the other two sensors). Unfortunately, many ML algorithms fail when faced with inference-time data incompleteness in the sense that their performance degrades drastically even when a small fraction of sensors fail.

Our research aims to design ML-based techniques that are robust against sensor/network failure-induced inference-time data incompleteness. Note that we do not intend to develop new ML algorithms; rather, we design novel techniques that can augment generic ML algorithms to enhance their robustness against inference-time data incompleteness. Several previous researchers have worked on overcoming sensor failures [15, 16, 17]. Most of these techniques and other existing related work are focused on providing efficient imputation algorithms to fill in the missing values of failed sensors with potential estimated values. These imputation-based strategies, however, are not effective when faced with simultaneous failures of multiple sensors (i.e., multiple data streams become simultaneously unavailable). This is because imputation-based strategies often suffer from the out-of-distribution (OOD) problem, where the new record produced by applying imputation deviates highly from the model's training data distribution, causing the model to misclassify such a record.

Towards enhancing the robustness of ML-coupled IoT systems against sensor failures, this paper presents SECOE – a novel approach to proactively alleviate sensor failure-induced inference-time data incompleteness. SECOE is fundamentally distinct from imputation-based approaches. The basic idea of SECOE is to create an ensemble [4] of ML models in which each model is specifically trained with an incomplete dataset representing a set of failed sensors. In other words, specific sensor feeds are omitted from training sets in anticipation of potential failures of corresponding sensors during inference time.

In designing SECOE, this paper makes the following unique technical contributions:

- We present the architecture of SECOE that includes novel techniques for ensemble creation and inference-time optimization, thereby demonstrating how the ensemble techniques can be harnessed for mitigating sensor failures in IoT systems.

- Towards improving training and inference efficiency, we present an algorithm that minimizes the sub-models in the ensemble. Our technique forms sensor groups based on inherent correlations among sensors in an IoT system. Furthermore, we present, Random-Selection, an alternate simple strategy for ensemble creation that randomly omits a set of sensors when training sub-models.

- We study the advantages and limitations of SECOE through extensive experimentation of our system in conjunction with three popular ML paradigms, namely, Multi-Layer Perceptron (MLP) [5], Random Forest (RF) [6] and Support-Vector Machine (SVM) [7]. We use three distinct datasets for our experimental study. Our study shows that SECOE is highly effective in overcoming sensor failure-induced data incompleteness.

The rest of the paper is organized as follows. The motivation and challenges are described in Section II, followed by Section III, in which we present the implementation details of SECOE. Section IV demonstrates the conducted experiments and results. Then Section V describes the related work. Finally, we conclude in Section VI.

II. MOTIVATION AND CHALLENGES

Due to its unquestionable capability of connecting a variety of physical objects using digital embedded objects, e.g., sensors, IoT has gained enormous attention and been recognized by a wide range of industrial organizations. Thus, several IoT applications have been developed in different domains, such as healthcare, smart environments (e.g., office and agriculture), and transportation and logistics. For instance, UPS and John Deere utilize IoT-enabled fleet tracking to improve supply efficiency and cut overall costs [8]. Another factor contributing to IoT's current prominence and widespread adoption across all industries is that major corporations are already investing billions in emerging technology. According to [9], billions of connected IoT devices, coupling the digital and physical worlds, are estimated to generate around $2.8-$6.3 trillion in potential IoT economic value by 2025.

The unique characteristics of IoT environments pose significant challenges for developing effective ML-based applications. Among these challenges, Incomplete data at inference time in IoT-ML-coupled systems is a critical challenge that this research aims to address. Inference time data incompleteness happens owing to data stream interruptions. In IoT systems, it is common for sensors to fail to send data to the IoT edge device/Cloud. This failure might be caused by delay, disconnection, and other hardware limitations (e.g., electronic malfunctioning or battery failure). In the face of such failures, ML applications are required to make inferences based on incomplete/partial data. ML models, however, need an input size "shape" similar to the size of the input on which they were trained. Consequently, if sensor data is insufficient, the IoT-ML-based system would be unable to process it.

Several earlier researchers, including [15] and [17], have tried to address the problem of data incompleteness caused by sensor failures. The majority of earlier research in this field mostly focused on providing imputation methods to replace missing data with estimated values. However, in the situation of simultaneous sensor failures, imputing their missing data may result in the OOD issue, resulting in severe performance degradation of the model. In contrast to past research, our objective is to optimize the performance of the IoT ML-based system in the presence of concurrent sensor failures by limiting the number of missing value imputations. Note that we do not propose to construct new ML algorithms; instead, we design novel techniques that can augment generic ML algorithms in order to improve their robustness against inference-time data incompleteness.

Our research objective is to proactively lessen the effects of missing data caused by sensor failures in IoT ML-based systems using an ensemble of a few sub-models, each of which is trained on a distinct subset of sensors chosen based on the correlation of sensors from the total IoT sensors. To the best of our knowledge, none of the existing work attempts to minimize imputing the missing data during the production time of the ML-IoT-based system. In this research, we introduce SECOE, an approach that enables the IoT ML-based system to make a real-time prediction without the necessity of filling in the missing value of the failed sensor when up to ~50% of the IoT sensors are failed. Hence, the IoT application could continually make a real-time prediction with an optimized accuracy using the correlated sensors to the failed ones, giving the IoT infrastructure's operator more time to physically inspect the failed sensors and replace them if necessary.

III. SENSOR CORRELATION-BASED ENSEMBLE (SECOE)

*A. Overview*

The main objective of the presented approach in this paper is to reduce the impact of sensor failures on the performance of IoT ML-based applications during inference time. Our method leverages an ensemble of a few sub-models each trained on a distinct subset of sensors formed based on the correlation of sensors in the IoT system. We use the correlation between sensors to minimize the number of models and build sub-models that perform similarly to the base model, which is a model trained using the entire IoT sensors data of the system.

Fig. 1 shows an overview of the training architecture of our method. Initially, our method groups the correlated sensors together, producing different groups of correlated sensors. Then, after computing the minimum number of sub-models (*MinM*), for instance, four sub-models, our method forms the features for each sub-model by selecting a distinct ~50% of sensors from each correlated group. The sub-model features created will then be used to build the sub-models along with the base model. The sensors included in each model's features correspond to ~50% of the total sensors of the IoT system. Our method guarantees that for each sensor x in the IoT system, there exists at least one suitable sub-model trained using a subset of sensors that includes some correlated sensors to sensor x, excluding sensor x. Thus, during inference-time, if a sensor x fails, then the prediction accuracy of the IoT application is maintained via an ensemble of the suitable sub-models or at least one of the sub-models.

The following subsections describe the architecture of our approach in more detail.

*B. Forming Correlated Groups*

The correlation between sensors is the key factor of our approach. We have only considered the positive Pearson correlation measurement from 0 to 1 scale when forming the groups of correlated sensors. We have categorized the coefficient interpretations into four categories, which are strong, moderate, weak, and very weak correlation. The coefficient correlation value ranges are 0.76-1.0, 0.50-0.75, 0.25-0.50, 0.10-0.25 for the above-mentioned categories, respectively. Each sensor is grouped with its highly correlated sensor. Note that a correlated group $G_i$ could contain correlated sensors with different coefficient interpretations.

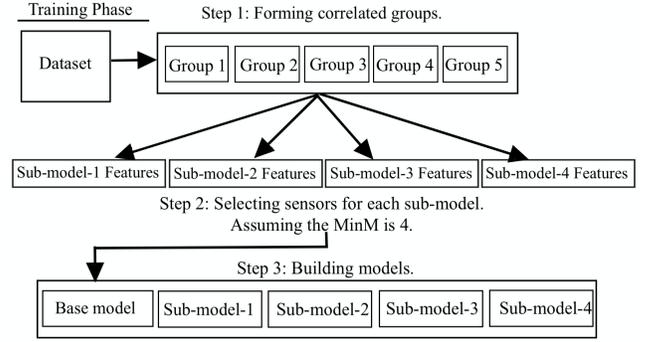

Fig. 1 Architecture of the training phase.

*C. Selecting Sensors and Building the Sub-models*

Selecting sensors for each sub-model is based on Algorithm 1. Each sub-model's features can have a different 50% of the sensors from each correlated group $G_i$. If the size of a $G_i$ is an odd number, then we select sensors for each sub-model equal to the nearest integer to 50% of the total sensors in such a $G_i$ (i.e., No. of sensors in $G_i$ is 7, then sensors that would be selected for each sub-model from such a $G_i$ is 4). For reliability, a sensor must be included in *at least* one sub-model and excluded from another sub-model.

As a result, when up to almost 50% of sensors from each correlated group $G_i$ fail, our proactive solution still provides a sub-model, trained on the remaining 50% of sensors (free-of-failure), that could make a real-time prediction with high accuracy using the correlated sensors to the ones that have failed. Furthermore, in case of the failure of many sensors

---

**Algorithm 1:** Sensors Selection For Sub-Models, SSMs

**input** : Correlated Groups, $CG = \{G_1, G_2, ..., G_n\}$, Size of The Largest Correlated Group, $L$
**output:** set of Models Features, $MF^* = \{MF_1, MF_2, ..., MF_n\}$

1 //Initialization
2 **if** $L$ *is odd* **then**
3 $\quad$ $MinM \leftarrow$ Round(0.50 * L)
4 **else**
5 $\quad$ $MinM \leftarrow$ Round(0.50 * L) +1
6 **end**
7 //List declaration for each model's features
8 **for** $n \leftarrow 1$ *to MinM* **do**
9 $\quad$ $MF_n \leftarrow []$
10 $\quad$ $MF^* \leftarrow$ Append($MF_n$)
11 **end**
12 **for** *each correlated group*, $G_i \in CG$ **do**
13 $\quad$ $H \leftarrow$ Round(0.50 * length($G_i$))
14 $\quad$ **for** *each model feature*, $MF_j \in MF^*$ **do**
15 $\quad\quad$ $MF_j \leftarrow$ Extend($G_i[sensor_1, .., sensor_H]$)
16 $\quad\quad$ $G_i \leftarrow$ SHIFT LEFT By 1($G_i$)
17 $\quad$ **end**
18 **end**
19 **return** ($MF^*$)

concurrently, leading to the necessity of imputing the missing values of such sensors, the number of imputations is minimized by at least ~50% less than the base model.

See Algorithm 1 for the pseudocode of sensors selection per sub-model. Algorithm 1 requires two arguments: lists of the correlated groups *CG* and the size of the largest correlated group *L*. First, Algorithm 1, using (1), finds the minimum number of the sub-models (*MinM*). Equation (1) finds *MinM* in the case of the percentage of sensors that would be selected from each $G_i$ is set to 50%. *MinM* satisfies these two conditions: (a) Each sensor from each $G_i$ is at least included in one sub-model, and (b) Each sensor is at least excluded from one sub-model.

$$MinM = \begin{cases} round(50\%*L)+1, & \text{if } L \text{ is even.} \\ round(50\%*L), & \text{otherwise.} \end{cases} \quad (1)$$

The above two conditions guarantee that if any sensor x from the IoT system fails, there is a sub-model, trained without sensor x, to make a real-time prediction.

Second, Algorithm 1, for each sub-model, selects the leftmost sensors from each $G_i$, equals to ~50% of sensors in $G_i$, then it shifts each $G_i$ to the left by 1. This step ensures that sensors that will be picked for the next sub-model differ from those selected for the previous sub-model. As illustrated in Fig. 2, if we have a correlated group $G_1$ as [H, B, D, C, A, G], Algorithm 1 will select the leftmost three sensors ([H, B, D]) from $G_1$ for the first sub-model, then shift $G_1$ to the left by 1. Next, for the second sub-model, Algorithm 1 will choose the leftmost three sensors ([B, D, C]) from $G_1$ after being shifted. This procedure is repeated for the following sub-models. Finally, after iterating over every $G_i$, Algorithm 1 outputs the model features $MF_j$ for each sub-model. The resulted features for each sub-model is approximately 50% of the entire IoT sensors.

### D. Accuracy Optimization

To optimize the performance of the IoT ML-based system during the presence of sensor failures, whenever records are received from sensors for prediction, they would be fed to the most suitable sub-model, the model that is trained on the subset of sensors that contains the least matched number of the failed sensors. The final prediction is their majority voting (ensemble) if there are several suitable sub-models. However, in some cases in which only two suitable models exist, the model with the highest training accuracy will be chosen for prediction. Note that the base model handles the prediction in the case of no failure of sensors. See Fig. 3.

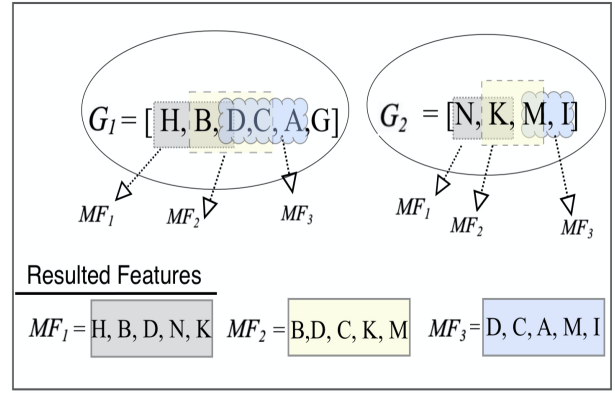

Fig. 2 Example of selection of sensors from two correlated groups.

## IV. EVALUATION

### A. Datasets

We evaluate our method on three datasets from the UCI repository [10]: (1) Dry-Beans, (2) Steel Plates Fault, and (3) Wall-Following Robot Navigation. Due to the lack of publicly available datasets containing features corresponding to sensors' readings with target class label/labels, we have utilized the first two datasets since they have a fair amount of correlation between features that contain either real or integer data types. We slightly preprocessed these two datasets. More specifically, we rescaled the features of each training dataset by subtracting the mean and dividing all values by the standard deviation of the training samples. The third dataset contains actual sensors' readings. Dry-Beans contains more than 13K samples forming 16 dry beans' geometric features to classify the bean into one of 7 species. In contrast, the Steel Plates Fault consists of 27 attributes describing the plates' geometric shapes and 1941 instances to classify Steel plates' faults into one of seven different types. The third dataset, Wall-Following Robot Navigation, comes with different versions having 2, 4, and 24 attributes corresponding to numerical ultrasound sensors' readings collected from sensors embedded on a SCITOS G5 robot navigating in a room following the wall in a clockwise direction. We used the version that contains 24 attributes. This version has more than 5K instances, through which the movement of the robot is predicted into one of 4 classes: move-forward, slight-right-turn, sharp-right turn, or slight-left-turn. For clarity, in our experiments, we renamed features of all datasets to follow the English alphabetical order from A-Z and a-z.

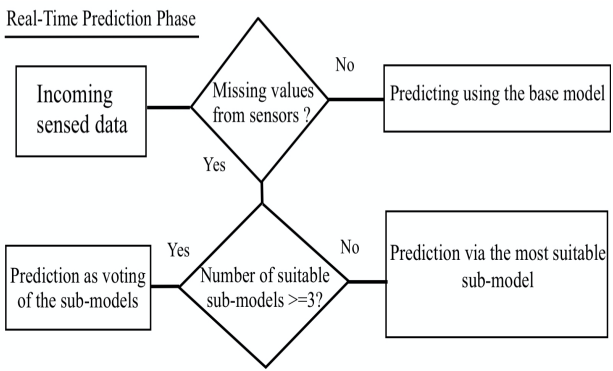

Fig. 3 Procedure of real-time prediction.

TABLE I. CORRELATED GROUPS AND SUB-MODELS' FEATURES.

| Dataset | Correlated group *G* | Sub-model features (MF) |
|---|---|---|
| Dry-Beans | $G_1$= {H, B, D, C, A, G}<br>$G_2$= {E, F}<br>$G_3$= {N, K, M, I, L, O}<br>$G_4$= {J, P} | $MF_1$= {H, B, D, E, N, K, M, J}<br>$MF_2$= {B, D, C, F, K, M, I, P}<br>$MF_3$= {D, C, A, E, M, I, L, J}<br>$MF_4$= {C, A, G, F, I, L, O, P} |

## B. Experimental Setup

Initially, we examine SECOE when all models are MLPs [5] trained on 85% of the dataset size and tested on the remaining 15%. Next, to investigate whether the training set size would affect the performance of our method, we conduct different experiments in which all ML models are trained on three portions {90%, 80%, 75%} of the dataset and tested on the remaining portions. To validate the generality of SECOE across different ML methods, we also examine its performance with two additional supervised ML algorithms: RF [6] and SVM [7]. In addition, we ran some experiments to find whether increasing the number of models above the *MinM* would affect the performance of our approach. To show the effectiveness of our SECOE, we compare SECOE with two other approaches. The first one is the base model. The second approach is Random-Selection, an approach we have created similar to ours. Instead of using correlation of sensors, Random-Selection randomly selects 50% of sensors from the entire sensors of the IoT system for each sub-model. We evaluate each approach based on its classification accuracy of the testing set. For simplicity, in all experiments, we utilize the Mean Imputation technique to replace the missing sensors' values with the Mean values of their records across the training set. All models are built using the default parameters defined by the "*Sciket-Learn*" python library [11].

## C. Performance of Models

In this subsection, we illustrate the performance of all models when there are no sensor failures. The *MinM* that SECOE produced is six sub-models for the Steel Plates Fault dataset and four for the other two datasets. Table 1 displays the correlated groups and sub-models' features resulting from Algorithm 1 on the Dry-Beans dataset. Fig. 4 demonstrates the classification accuracy of the sub-models of both SECOE and Random-Selection and the base model on the three datasets: Dry-Beans, Steel Plates Fault, and Wall-Following Robot Navigation. In Fig. 4, all models are MLPs, trained on 85% and tested on the remaining 15% of the dataset. Although they are built using ~50% of the sensors from the dataset, most of the sub-models by SECOE are comparable to the base model in terms of test accuracy. In contrast, some Random-Selection sub-models perform poorly; more specifically, sub-model-4 and sub-model-1 on Steel Plates Fault and Wall-Following Robot Navigation datasets, respectively.

Choosing from the dataset a subset of sensors that contains the important sensors, which are the sensors that have a high impact on the model classification accuracy, is essential to make the ML model performs well. Since Random-Selection randomly selects sensors for each sub-model, some important sensors or their correlated ones were ignored (not included in any sub-model), producing weak-performed models. SECOE offers well-performed sub-models built using the important sensors or ~50% of its correlated ones. Overall, SECOE provides better-performed sub-models than the Random-Selection approach.

## D. Static Sensor Failure Evaluation

To show the importance of SECOE, we created several test cases in which specific sensors were statically chosen as the simulated failed sensors. Each test case corresponds to a different percentage of simultaneous sensor failures, from 5% up to 50%. In this experiment, for each test case, there are suitable sub-models or at least one suitable sub-model, free of sensors' failure, trained on the correlated sensors to the failed sensors. Fig. 5 demonstrates the results of the test case scenarios for each dataset. As can be seen in Fig. 5, the test accuracy of the base model drops sharply when the percentage of failed sensors is 30% or more. At the highest percentage of sensor failures, SECOE maintains the accuracy at 90.84%, 73.63%, and 84.98%, which are 77.49%, 61.64%, and 94.42% more than what has been achieved by the base model on the Dry-Beans, Steel Plates Fault, and Wall-Following Robot Navigation datasets, respectively. Such a dramatic drop in the base model's accuracy is due to the OOD problem caused by imputing many missing values of the failed sensors. Substantially, SECOE minimizes the percentage of imputation to 0% compared to the base model. These results confirm the intuition behind SECOE: by using a few sub-models built carefully using correlation of sensors, we can achieve high classification accuracy even though up to 50% of sensors fail without the necessity of imputing the missing values of these sensors.

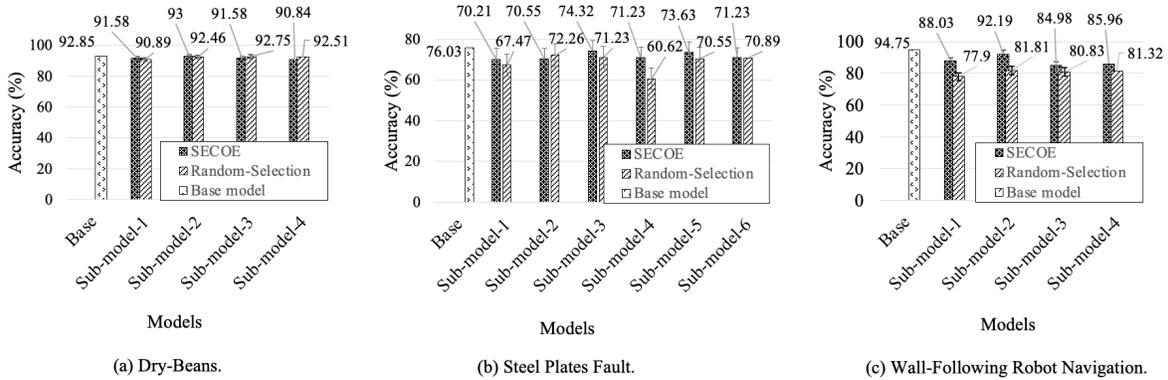

Fig. 4 Test accuracy comparison between the base model, Random-Selection and SECOE sub-models on each dataset.

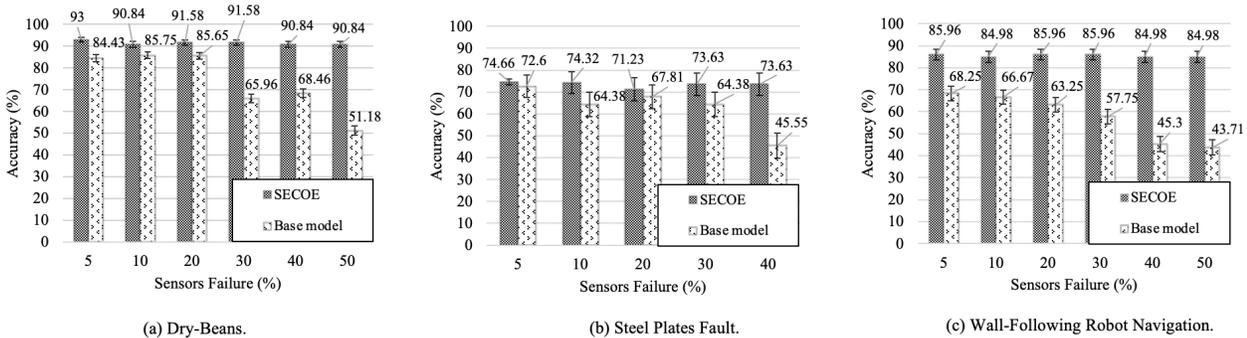

Fig. 5 Test accuracy with a 95% confidence interval of SECOE versus the base model during the failure of sensors on each dataset.

### E. Random Sensor Failure Evaluation

To precisely examine our approach, we have mimicked real-world scenarios by simulating several test cases where sensors failed concurrently at random. We compare our SECOE with the Random-Selection and base model approaches under seven different fractions of failed sensors, ranging from 5% to 60%. Fig. 6 shows, for all datasets, the average test accuracy of 10 iterations per test case. The results indicate that SECOE outperforms the base model and Random-Selection. When less than 30% of sensors fail on the Dry-Beans and Steel Plates Fault datasets, the base model exhibits performance close to our SECOE. However, its performance decreases severely when the failure rate of sensors is above 30% on all datasets. More specifically, when 60% of sensors fail, the accuracy of the base model reaches 56.83%, 42.43%, and 51.79% on the Dry-Beans, Steel Plates Fault, and Wall-Following Robot Navigation datasets, respectively. These results are 13.10%, 14.14%, and 9.57% less than what our SECOE has obtained on the three datasets, respectively. In SECOE, since each sub-model is trained on a distinct ~50% of the entire IoT sensors from the dataset, it minimizes the number of imputations of the missing sensors' values by at least ~50% less than the base model. Therefore, when more than 30% of the sensors in the system fail, the ensemble of sub-models by SECOE obtains better classification accuracy than the base model. On the other hand, due to its random selection of sensors, Random-Selection gets the lowest average accuracy in every simulated test case across all datasets, except the Wall-Following Robot Navigation. It has similar performance to the base model when the percentage of failed sensors is 30% or above.

From analyzing these results, we can draw the conclusion that using the correlation of sensors to build an ensemble of sub-models alleviates the impact of concurrent sensor failures. To further enhance the performance of SECOE, we suggest that future research should investigate the performance of SECOE utilizing alternative imputation algorithms that take sensor "features" correlation into account when calculating the missing value of a sensor.

### F. Ablation Studies

To further validate SECOE, we conduct similar experiments to those in the previous subsection IV.E using different training and testing set sizes. Fig. 7 shows the results of these experiments on the Dry-Beans dataset. These results align precisely with the prior experimental results presented in Fig. 6 on the same dataset, implying that the tested training sizes do not influence the performance of SECOE and other approaches.

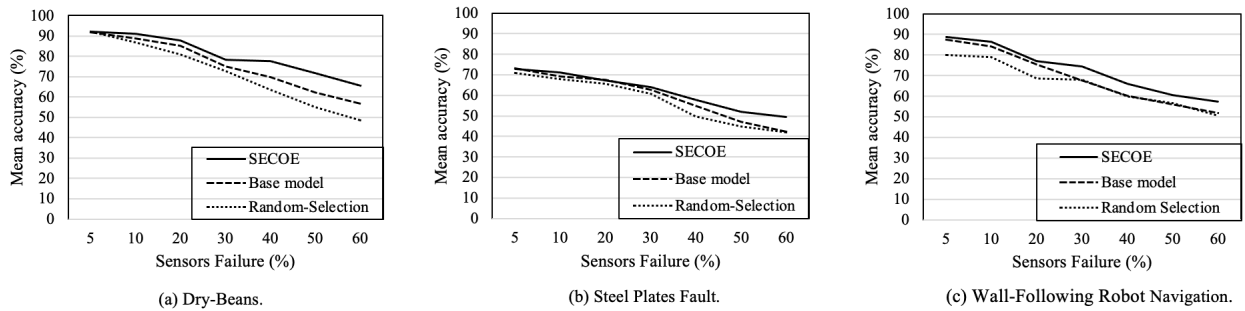

Fig. 6 Mean test Accuracy of SECOE, Random-Selection, and Base model during random failure of sensors on each dataset.

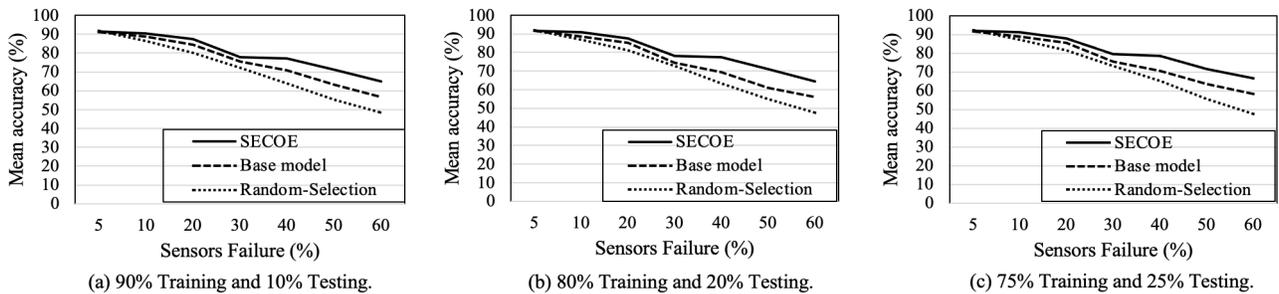

Fig. 7 Mean test Accuracy comparison on the Dry-Beans dataset when the models are trained on different training portions.

We do not show figures of the experimental results on the other two datasets since they are also very similar and consistent with the results shown in Fig. 6.

On the other hand, to determine whether the type of ML algorithm influences the performance of SECOE, we have run the same experiments described in subsection IV.E using SVM and RF. The experimental results on the Dry-Beans and Steel Plates Fault datasets are plotted in Fig. 8 and Fig. 9, respectively. According to Fig. 8 and 9, although the supervised ML methods RF, MLP, and SVM vary in terms of learning, the performance of SECOE using SVM and RF is consistent with its results using the MLP ML method and better compared to the base model and Random-Selection. These results prove the generality of SECOE across different ML algorithms. In the case of RFs, the Random-Selection outperforms the base model as concurrent sensor failures increase. This is because RF is an ensemble of decision trees. Hence, having several RF sub-models improved Random-Selection's performance over the base model.

*G. Effect of The Number of Models*

Algorithm 1 finds the *MinM* that, for reliability, ensures that each sensor from each correlated group $G_i$ is at least included in one sub-model and omitted from one sub-model. The question is: how would the performance of SECOE be affected when we increase the number of models above the *MinM*? To answer this question, we have run SECOE on the Wall-Following Robot Navigation dataset when the number of models is 4, 6, 12, and 14, where 4 is the *MinM*. Fig. 10 shows the effect of the number of models on the performance of SECOE. According to Fig. 10, the performance of SECOE is better when we increase the number of models over the *MinM*. Although 6 is half of 12, it has shown quite similar performance compared to the 12 and 14 models, except when 40% of sensors failed. This performance improvement is because SECOE uses majority voting of the most suitable sub-models (ensemble of sub-models). Therefore, using a larger number of models than the *MinM* increases the number of most suitable sub-models, optimizing the prediction accuracy. These results infer that increasing the number of models slightly larger than the *MinM* is always recommended for better performance.

V. RELATED WORK

Several previous works aimed to address data incompleteness. Some of which utilize probabilistic matrix factorization (PMF) to recover missing sensors' value(s). As a recent example, Fekade *et al.* proposed a method that recovers the missing values of a sensor by performing PMF to its group of related sensors, which has been preliminarily assigned using the K-means clustering algorithm. Their method has shown better performance than SVM and deep neural network (DNN) algorithms [12].

Neural networks (NN) have also been used for building efficient imputations to fill the missing values. Wang *et al.* offered a gated recurrent unit filling (GRUF) at the edge nodes for artificial IoT (AIoT) applications. Their method, based on the GRU and generative adversarial network (GAN) algorithms, effectively fills the missing value(s) of sensors by learning the internal properties of IoT data via sensor geographic location and time series data history. Empirical results have proven that GRUF outperforms all comparable imputation techniques, such as the Mean [13].

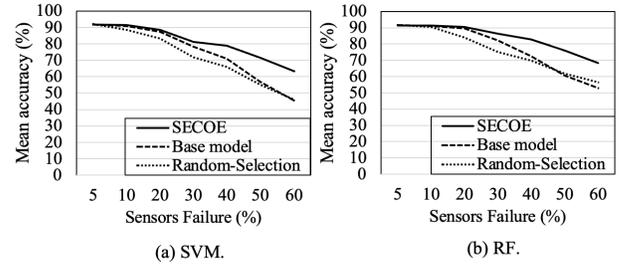

(a) SVM.    (b) RF.

Fig. 8 Performance of SECOE using SVM and RF on the Dry-Beans dataset.

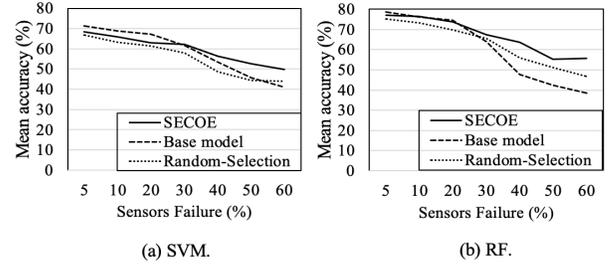

(a) SVM.    (b) RF.

Fig. 9 Performance of SECOE using SVM and RF on the Steel Plates Faults dataset.

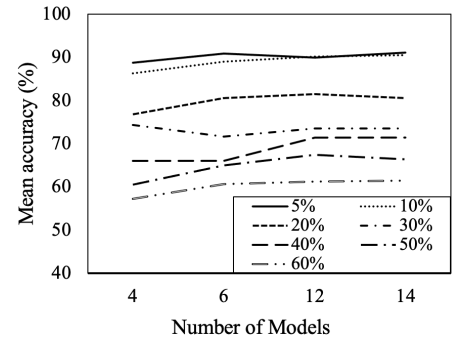

Fig. 10 Comparison of SECOE on a different number of models at different sensors failure percentages on the Wall-Following Robot Navigation dataset.

Using the correlation of features helps in constructing efficient imputation methods. Mary and Arockiam proposed ST-correlated, an imputation approach that uses the temporal correlation of IoT sensors to proximate sensor readings corresponding to time [14]. At different percentages [5%, 10%, and 15%] of missing values, their ST-correlated outperforms different imputation methods, namely Mean, Median, Mode, and MICE from R. Ilyas *et al.* presented a framework to mitigate sensors' failure [15]. Their framework finds the highly correlated IoT sensors in the surrounding environment of any type, not within the same IoT system. Then, based on the metadata of the nearby sensor, such as location and sensor type, their framework creates an ML virtual sensor using the historical data of the faulty and the nearby correlated sensors. Tsai *et al.* proposed real-time anomaly detection and recovery in IoT sensing systems [16]. Their method constructs a Bayesian model for each sensor utilizing its Top-2 linked sensors. This Bayesian model predicts sensor data. After analyzing defective patterns, they provided a technique that changes sensed data based on the kind of fault to make it look normal.

Unlike all the previous works, Fonollosa *et al.* offered sensor failure mitigation using several SVM models [17]. The authors employ the mathematical combination formula $C(n, r)$ to build different sub-models, where $n$ is the total number of sensors, and $r$ represents the number of faulty sensors. Sub-models built with the same number of sensors are grouped as a classifier.

Among the above-related works, only [15, 16, 17] dealt with missing data during the in-production of the application. [15] relies on accessing nearby sensors' data from different IoT systems. Such access requires permission, which is not always granted. [16] does not address complete sensor failure (complete missing sensor values). In general, all the above methods mainly focus on providing data imputation techniques to fill in missing values, except [17], which employs many models and uses zero imputation to alleviate data incompleteness. In this paper, we focus on providing a method that optimizes the prediction accuracy of the IoT ML-based system by minimizing the imputation of missing sensors' values.

## VI. Conclusion

Due to its properties, IoT faces several fundamental challenges in designing and deploying efficient ML IoT applications. One such challenge is sensor/network failures that result in data stream interruptions. Unfortunately, in the presence of missing data, many ML systems encounter rapid degradation in their performance. In this paper, to make IoT-coupled ML systems robust against simultaneous sensor failures, we introduced SECOE, an ensemble of sub-models, each constructed utilizing distinct subsets of sensors obtained through a method that employs sensor correlation. In addition, we created a Random-Selection technique for comparison purposes. A series of empirical studies carried out have proven the intuition behind SECOE. At various percentages of simulated concurrent missing sensor data streams, SECOE enabled the IoT ML-based system to produce accurate real-time predictions without having to fill in the missing sensor readings. When 40-50% of sensors simultaneously failed, SECOE produced classification accuracies that were substantially better than the base model by 77.49%, 61.64%, and 94.42% on the three datasets, respectively. In other scenarios in which SECOE was examined when sensors were failing randomly and imputation of their values was required, SECOE achieved a noticeably higher classification accuracy than the base model and Random-Selection. Furthermore, SECOE reduced the imputation of missing sensor data streams by at least 50% less than the base model. We also studied the performance of SECOE with varying numbers of sub-models and provided a suggestion for the number of sub-models for obtaining better performance of SECOE.

## Acknowledgment

This research has been partially funded by grants from the National Science Foundation (NSF) and the US Department of Agriculture's National Institute of Food and Agriculture (USDA NIFA), and gifts from Accenture Research Labs. Any opinions, findings, and conclusions or recommendations expressed in this material are those of the authors and do not necessarily reflect the views of the funding agencies or the companies mentioned above.